\documentclass[letterpaper, 10 pt, conference]{ieeeconf}  

\IEEEoverridecommandlockouts                              

\usepackage{cite}
\usepackage{graphics} 
\usepackage{epsfig} 
\usepackage{mathptmx} 
\usepackage{times} 
\usepackage{amsmath} 
\usepackage{amssymb}  
\usepackage{booktabs}
\usepackage{multirow} 

\title{\LARGE \bf
A VLM-based Method for Visual Anomaly Detection in Robotic Scientific Laboratories
}

\author{Shiwei Lin, Chenxu Wang, Xiaozhen Ding, Yi Wang, Boyuan Du, Lei Song,\\Chenggang Wang, and Huaping Liu%
\thanks{*This work was supported by the National Natural Science Fund for Key International Collaboration under grant 62120106005.}%
\thanks{Shiwei Lin, Chenxu Wang, Yi Wang, and Huaping Liu are with the Department of Computer Science and Technology, Tsinghua University, Beijing 100084, China. linsw23@mails.tsinghua.edu.cn, hpliu@tsinghua.edu.cn.}%
\thanks{Xiaozhen Ding is with the School of Physics and Electronic Information, Yantai University, Yantai 264005, China.}%
\thanks{Boyuan Du is with the School of Computer and Big Data, Fuzhou University, Fuzhou 350108, China.}%
\thanks{Lei Song and Chenggang Wang are with Department of Automation, Shanghai Jiao Tong University, Shanghai, 200240, P. R. China.}
\thanks{Huaping Liu is the corresponding author.}%
}

\begin{document}

\maketitle
\thispagestyle{empty}
\pagestyle{empty}

\begin{abstract}

In robot scientific laboratories, visual anomaly detection is important for the timely identification and resolution of potential faults or deviations. It has become a key factor in ensuring the stability and safety of experimental processes. 
To address this challenge, this paper proposes a VLM-based visual reasoning approach that supports different levels of supervision through four progressively informative prompt configurations.
To systematically evaluate its effectiveness, we construct a visual benchmark tailored for process anomaly detection in scientific workflows. Experiments on two representative vision-language models show that detection accuracy improves as more contextual information is provided, confirming the effectiveness and adaptability of the proposed reasoning approach for process anomaly detection in scientific workflows. Furthermore, real-world validations at selected experimental steps confirm that first-person visual observation can effectively identify process-level anomalies. This work provides both a data-driven foundation and an evaluation framework for vision anomaly detection in scientific experiment workflows.

\end{abstract}

\section{INTRODUCTION}

In scientific experiments, when unexpected events occur during the process, accurately and efficiently detecting anomalies through visual information is critical for dynamic task scheduling\cite{ouelhadj2009survey}, embodied perceptual decision-making\cite{liu2025embodied}, minimizing downtime, improving system efficiency, and reducing human intervention. It also contributes to the development of robust laboratory safety protocols~\cite{tom2024self}.
Most existing experimental designs in scientific laboratories focus on workflow planning and schedule optimization, while paying limited attention to potential anomalies that may arise during the experimental process\cite{darvish2025organa, yang2025human, boiko2023autonomous}. 
Current vision-based anomaly detection methods are mainly developed for specific domains such as manufacturing processes and abnormal behavior analysis \cite{sarker2023detecting, yan2022real, li2021decoupled, liu2023amp, zhang2022anomaly, yao2024prior, lukin2023anomaly, zhu2023minigpt, li2023myriad}, are not well suited to handle process anomaly detection in diverse scientific experiments, and exhibit limited transferability between diverse experimental setups.

\begin{figure}[t]
\centering
\includegraphics[width=\linewidth]{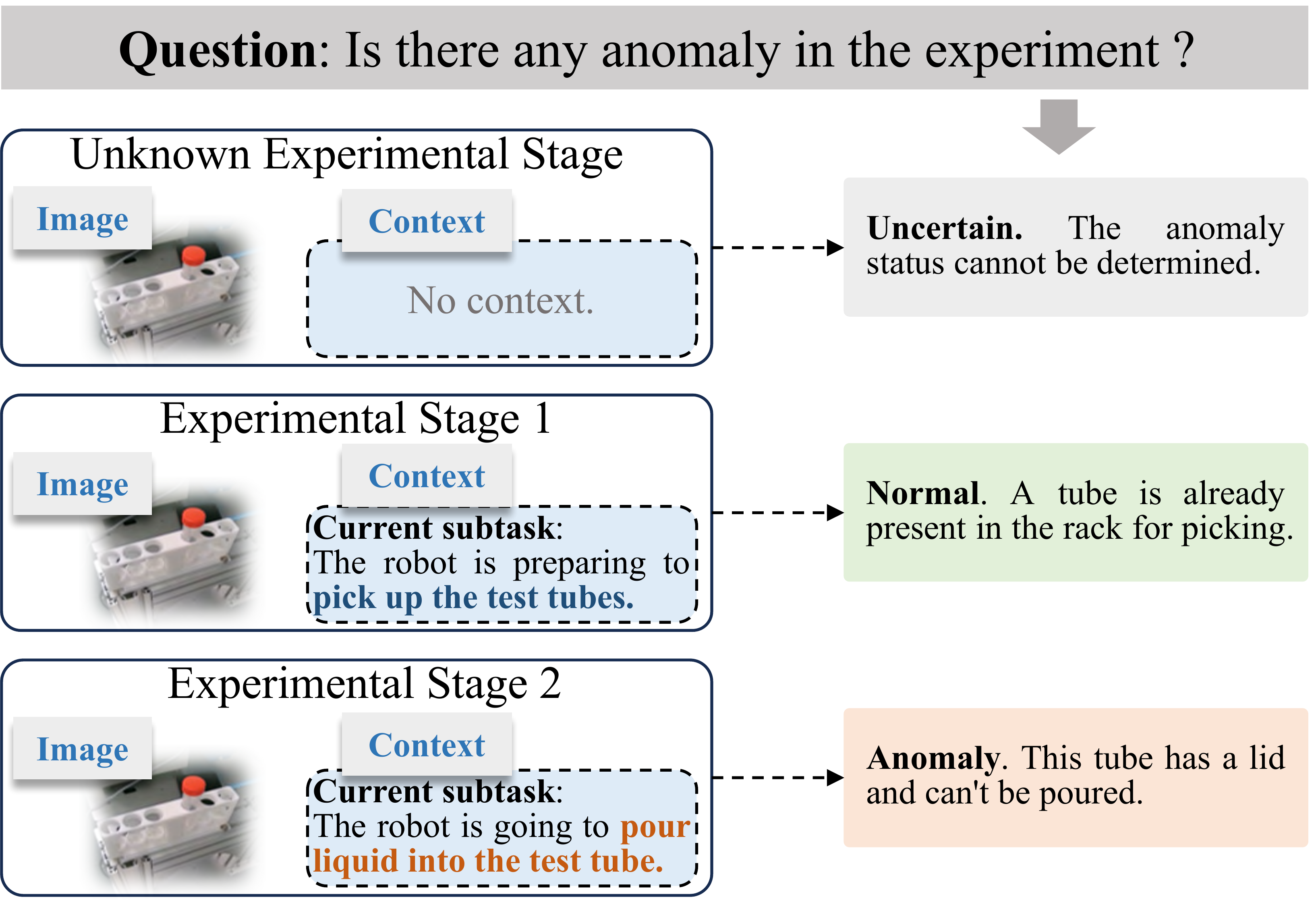}
\caption{Illustration of context-dependent anomaly determination. Each experimental stage is represented by a combination of a visual state (image) and a corresponding textual description (context). The same image may be considered normal or anomalous depending on the subtask context. (a) Without contextual information, the status of the image is ambiguous. (b) In a stage where the presence of the test tube is expected, the image is considered normal. (c) During the pouring stage, the absence of the test tube makes the image is considered abnormal.}
\label{fig:intro1}
\end{figure}

In scientific experiments, process anomaly detection is context-dependent, as the same target state may be normal or abnormal at different workflow stages. This context-sensitive nature makes it difficult for conventional classification-based methods to accurately identify such anomalies, as illustrated in Fig.\ref{fig:intro1}.

\begin{figure}[t]
\centering
\includegraphics[width=\linewidth]{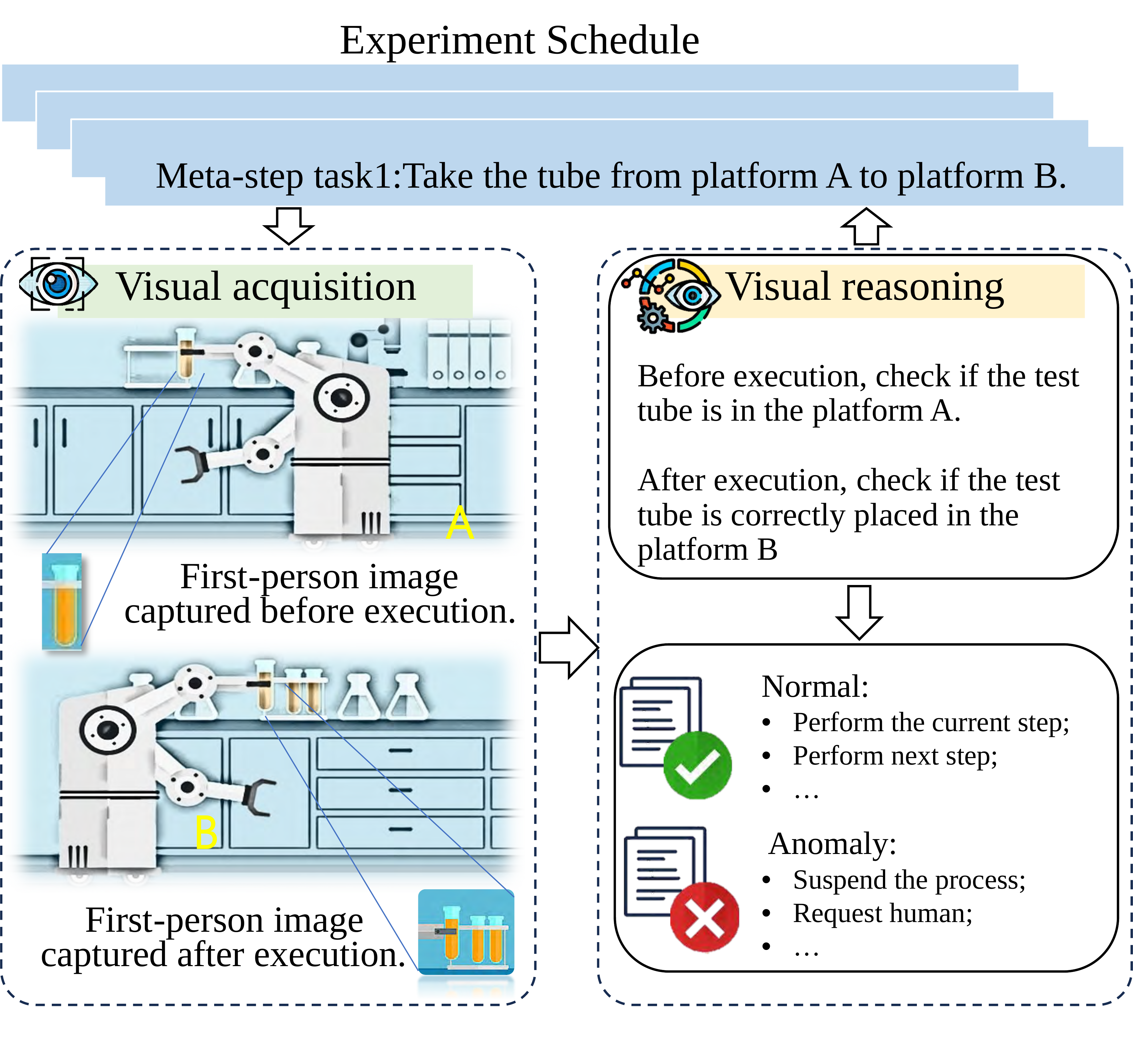}
\caption{Illustration of the process anomaly detection problem in scientific experiments. A robotic system performs experimental steps while collecting visual data from a first-person perspective. At each step, the visual observations are used to determine whether the current process state is normal or anomalous, based on contextual information from the experimental workflow. The visual detection results can influence the subsequent experimental schedule.}
  \label{fig:intro2}
\end{figure}

Recent advances have shown that VLMs possess strong multimodal reasoning capabilities, supporting zero-shot anomaly detection via context-aware VQA\cite{lukin2023anomaly,tan2023knowledge}, precise anomaly localization by integrating traditional industrial knowledge with visual encoders\cite{zhu2023minigpt, li2023myriad}, and enhanced visual tracking through linguistic guidance\cite{chen2024mamtrack,liu2022embodied}. These developments build on the growing role of LLM-powered agents in scientific domains\cite{wang2024survey, darvish2025organa, frey2023neural}. Therefore, we adopt VLMs for anomaly detection, as they are capable of reasoning about the semantic meaning of target states in images, making them more suitable for context-aware decision-making in experimental workflows. Specifically, we propose a method that utilizes visual information collected during experimental execution, combined with structured prompt information to perform anomaly detection using VLMs. As shown in Fig.\ref{fig:intro2}, this approach integrates experiment perception, visual-language reasoning, and contextual understanding to determine whether the current stage is anomalous.

To systematically evaluate the proposed method, we construct a multimodal visual benchmark for process anomaly detection in scientific experiments. Using the silicone preparation workflow, we collect and annotate first-person visual data in a real chemical laboratory, covering 15 key stages and 20 monitoring points. This benchmark supports both stage-level and full-process anomaly detection tasks. Based on this dataset, we introduce four progressively informative prompt configurations to study the impact of prompt granularity on VLM reasoning performance, enabling evaluation and analysis of different models.
We conduct systematic experiments on two representative VLMs, GPT-4o and Qwen2.5-VL-72B-Instruct, and demonstrate that prompt design plays a crucial role in improving model accuracy and decision robustness.

Our main contributions are summarized as follows:
\begin{itemize}
    \item We propose a VLM-based visual anomaly detection method that utilizes first-person images and stage-dependent semantics across diverse scientific scenarios, enabling systematic investigation of prompt granularity effects and providing an evaluation standard for process anomaly detection in scientific experiments.
    \item We construct a vision-based benchmark for process anomaly detection in scientific experiments, grounded in a real silicone preparation workflow in a chemical lab. The benchmark includes multi-step, multi-view first-person images with detailed annotations, supporting multimodal models with contextual input, enabling research on anomaly understanding in scientific workflows.
    \item We conduct real-world experiments to validate the effectiveness of our method in the selected stage, demonstrating the practical applicability of first-person visual anomaly detection.
\end{itemize}
\section{Related Work}

\subsection{Anomaly Detection in Automated Systems}

Anomaly detection has long been applied in manufacturing, behavior monitoring, and defect inspection. Traditional approaches based on statistical analysis and handcrafted features remain prevalent. For example, liquid motion patterns in video frames have been used to improve experimental safety~\cite{sarker2023detecting}. However, these methods struggle with adaptability in complex or dynamic environments. Spatiotemporal regression has also been applied for the detection of localized overheating in additive manufacturing~\cite{yan2022real}, but its computational cost limits the practical deployment.

Deep learning enables more effective feature extraction in industrial anomaly detection. Unsupervised methods improve the detection of human and vehicle anomalies in surveillance settings\cite{li2021decoupled}, though high false positives remain under changing conditions. AMP-Net, which fuses spatial and temporal features, performs well in complex scenarios\cite{liu2023amp}, but like many deep models, it relies on domain-specific training data and lacks generalization to diverse environments.

GANs with attention mechanisms have achieved high accuracy in surface defect detection~\cite{zhang2022anomaly}, but are sensitive to data distribution and resource-intensive. Transformer-based models that incorporate prior knowledge handle various types of anomalies\cite{yao2024prior}, yet require extensive annotations and high inference costs. Overall, while deep learning outperforms traditional methods, it still faces limitations in robustness, adaptability, and resource demands.

Despite progress in industrial contexts, existing methods often fail to generalize to the complex and dynamic workflows of scientific experiments.

\subsection{VLMs for Anomaly Detection}

With the rapid development of multimodal learning, Vision Question Answering (VQA) has become a flexible and extensible solution for anomaly detection. AnyAnomaly\cite{lukin2023anomaly} introduces a context-aware VQA framework that leverages user-defined textual anomaly descriptions to enable fast, zero-shot detection, evaluated on the proposed C-VAD dataset. MiniGPT-4 and Myriad enhance multimodal models by integrating prior knowledge from traditional industrial detection models with visual encoders, improving the accuracy of anomaly localization\cite{zhu2023minigpt, li2023myriad}, with MiniGPT-4 evaluated on the COCO caption benchmark and Myriad validated across MVTec-AD, VisA, and PCB Bank datasets. However, these methods are primarily designed for static production environments or fixed defect types, making them less suitable for the diverse processes and irregular events encountered in scientific experiments.

In video monitoring and tracking, adding textual descriptions can enhance visual features and improve behavior recognition\cite{chen2024mamtrack}. Applied to experimental settings, keyframes from different time points can be described in natural language and semantically compared using large models\cite{de2024large}, enabling the detection of both environmental and procedural anomalies. Additionally, few-shot learning improves model adaptability to unseen scenarios and rare anomaly types, offering a promising direction for low-resource conditions\cite{wang2020generalizing}.
Existing vision-based anomaly detection methods are primarily developed for fixed industrial scenarios and often fail to generalize to the complex and variable workflows of scientific experiments. This limitation mainly stems from their dependence on task-specific data and limited adaptability to diverse procedural structures.


\begin{figure}[t]
\centering
\includegraphics[width=\linewidth]{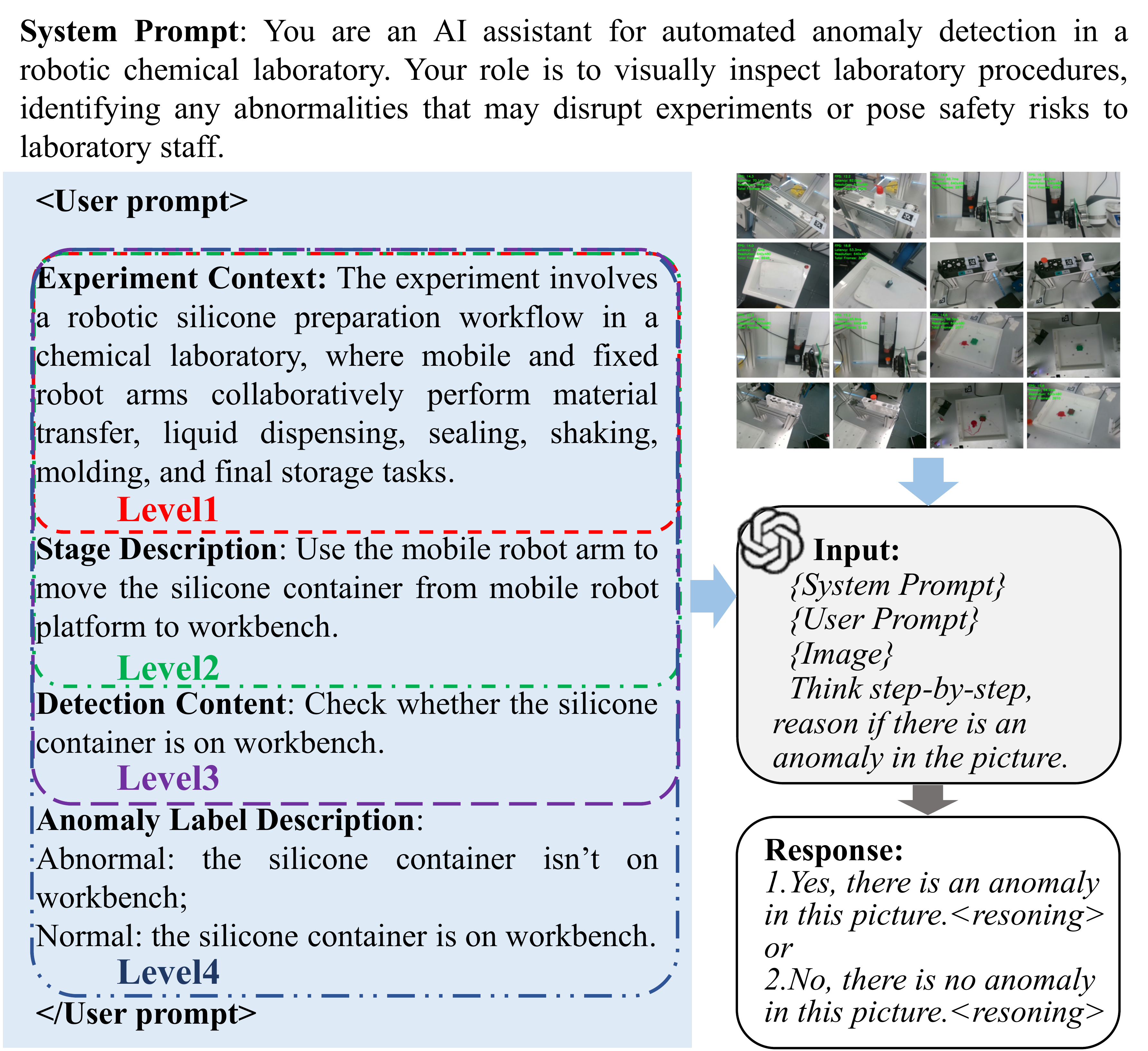}
\caption{Hierarchical prompt design and reasoning process. The left part shows the construction of prompts at different information levels, including experimental context, stage description, detection content, and anomaly label description. The right part shows how the VLM takes the image and prompt as input and performs step-by-step reasoning to identify potential anomalies.}
  \label{fig:context_set}
\end{figure}
\section{Problem Formulation}
A complete automated experimental workflow, as illustrated in Fig.\ref{fig:intro2}, can be formally represented by an ordered set of steps \( S = \{s_1, s_2, \dots, s_n\} \) along with fundamental information \( I \) about the context of the experiment. Each step \( s_i \), defined as a meta-step, may include one or multiple robotic actions. Each meta-step \( s_i \) requires certain prerequisites (e.g., materials, reagents, or equipment) and completion conditions, which together serve as the anomaly detection target \( C \), and may include a fine-grained anomaly description \( C_d \). 
To support different levels of prompt granularity in anomaly detection, we introduce an information control function \( \phi(S, C, C_d) \), which determines the semantic content provided to the model during inference. Given a visual observation \( x \) captured from the experiment, the anomaly detection task is formulated as a binary classification function:

\begin{equation}
\text{judgeAnomaly}(x, I, \phi(S, C, C_d)) \rightarrow y
\end{equation}

where \( y \in \{0, 1\} \) indicates whether an anomaly is present (1 for anomaly, 0 for normal). The function \( \phi \) selectively includes the step information \( S \), the detection target \( C \), and the fine-grained anomaly description \( C_d \), depending on the evaluation configuration. Formally, \( \phi \) serves as a semantic controller that determines the composition and granularity of the prompt content. One specific instantiation of \( \phi \) is presented in Section IV to support hierarchical prompt analysis.

This formulation provides a unified framework for modeling open-ended reasoning and goal-conditioned anomaly detection, and serves as the foundation for our benchmark's layered evaluation strategy.

\section{METHOD}
Our method takes a first-person image and structured textual information as input, and uses VLMs to perform multimodal reasoning. By progressively incorporating contextual and semantic information through prompt design, the model determines whether the current experimental state is normal or anomalous.

At each monitoring point within the experimental procedure, a first-person image is captured by the robotic system. This image is then paired with a natural language prompt that encodes progressively detailed contextual information. The prompt may include the experimental background, the specific subtask being performed, the detection target, and a semantic description of abnormal conditions.

The resulting multimodal input is fed into a VLM, which jointly processes the visual and textual information to infer the state of the experiment. Rather than producing a direct binary classification, we adopt a reasoning-based approach using Chain-of-Thought (CoT) prompting, which guides the model to analyze the situation step by step before reaching a conclusion. The final output is then used to assess whether an anomaly is present and to support subsequent experimental decision-making.

\subsection{Multimodal Input Formatting}

To enable effective reasoning by VLMs, our method formats both visual and textual information into a unified multimodal input. This section describes how the two modalities—first-person images and structured prompts—are prepared and combined, as illustrated in Fig.\ref{fig:context_set}.

\textbf{Visual Input}. The visual input consists of RGB images captured from a first-person perspective by robotic arms during the execution of each experimental stage. These images provide contextual views of critical equipment, materials, and objects in the workspace. All images are resized to 640$\times$480 resolution before being passed into the model to match input constraints and maintain visual consistency.

\textbf{Prompt Structure}. The textual input follows a structured natural language format and is dynamically generated based on the current experimental stage and prompt configuration level. Each prompt is composed of up to four textual elements: Experiment Context, Stage Description, Detection Content, and Anomaly Label Description. 
\begin{itemize}
    \item \textbf{Experiment Context:} A brief description of the overall experimental background.
    \item \textbf{Stage Description:} A description of the current subtask, includes the information about the operator, target object, start and destination position, and actions.
    \item \textbf{Detection Content:} A structured sentence that specifies the content to be checked in the scene, typically in the form of “Check whether [object] is [in a specific state or location]”.
    \item \textbf{Anomaly Label Description:} A semantic-level textual description that defines the condition under which a state is considered abnormal or normal.
\end{itemize}

\textbf{Multimodal Integration}. The image and prompt are submitted as a combined query to VLMs via its multimodal input interface. No additional segmentation tokens are required; the model handles joint encoding of vision and text natively. This consistent formatting ensures that the model receives clearly organized information, allowing it to focus on relevant aspects of the visual scene during reasoning.

\subsection{Hierarchical Prompt Design}

To investigate how the granularity of textual input affects model reasoning, we design a four-level hierarchical prompting scheme. Each level incrementally adds semantic or contextual information, enabling progressively more detailed guidance for the model.

\begin{itemize}
    \item \textbf{Level 1}: Contains only the \textit{Experiment Context}, which provides a general background of the experiment and helps the model understand the overall task setting.
    \item \textbf{Level 2}: Includes both the \textit{Experiment Context} and the \textit{Stage Description}, describing the specific subtask being performed in the current stage.
    \item \textbf{Level 3}: Adds the \textit{Detection Content} on top of Level 2, specifying what should be visually checked in the current image.
    \item \textbf{Level 4}: Further incorporates the \textit{Anomaly Label Description}, offering a semantic-level explanation of how normal and abnormal states differ.
\end{itemize}

This layered design enables a systematic evaluation of how varying prompt specificity influences model behavior. A visual illustration of this hierarchical prompt structure is provided in Fig.\ref{fig:context_set}. It also directly corresponds to the instantiation of the prompt function $\phi_i$ defined in our problem formulation, where each level represents an increasing degree of semantic richness and contextual precision in the input.

This progressive design allows us to systematically evaluate the effect of prompt granularity on model performance. As more context and guidance are provided, the model is expected to exhibit more accurate and consistent reasoning capabilities.

\subsection{Reasoning Process}

Given a first-person image and its corresponding hierarchical prompt, the VLM is required to jointly interpret both the visual input and structured textual information to assess whether the current experimental state is normal or anomalous. Rather than performing direct classification, the model is guided to reason step by step through Chain-of-Thought (CoT) prompting. This approach encourages more interpretable and robust decisions, especially under complex or ambiguous visual conditions.

The model output is a natural language response that typically includes reasoning steps followed by a conclusion. To derive the final anomaly label, we apply a post-processing procedure that converts the model's conclusion into a binary decision. Specifically, a rule-based natural language parser is employed to extract the final decision from the response.

This reasoning and decision pipeline ensures that model predictions are semantically grounded and suitable for comparison against annotated ground truth labels in the benchmark.

\begin{figure}[t]
  \centering
  \includegraphics[width=\linewidth]{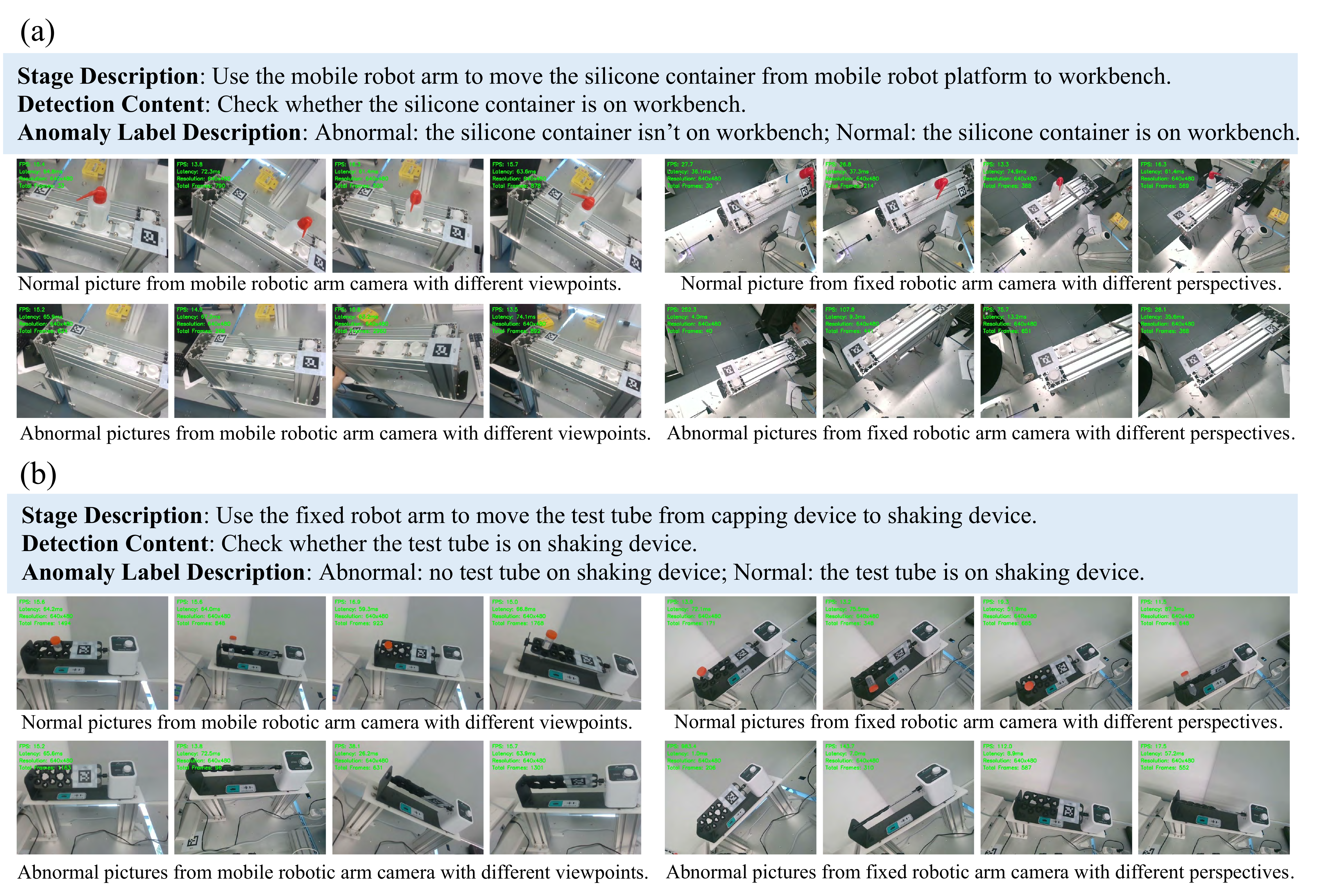}
  \caption{Examples from two detection points in the proposed benchmark. Each example includes first-person images captured from different devices and viewpoints, along with their associated textual annotations. (a) Images captured near the workbench, show both the presence and absence of the silicone container on the table. (b) Images captured near the shaking device, show both the presence or missing on the shaking device.}
  \label{fig:pic_set}
\end{figure}
\section{Benchmark Implementation}
The benchmark is constructed based on a real chemical laboratory setting. It provides a unified dataset, task definition, and evaluation protocol to systematically assess the performance of multimodal models in detecting anomalies during experiments.

Our dataset captures a complete automated silicone preparation process and contains 1001 first-person images, including 501 normal and 500 abnormal samples. The workflow is divided into 15 discrete stages, with 20 monitoring points. Visual data are collected using both fixed and mobile robotic arms from multiple viewpoints to ensure diversity in spatial and visual configurations. Examples are shown in Fig.\ref{fig:pic_set}.

Each image is accompanied by structured textual annotations that provide contextual information for multimodal reasoning. These annotations include four distinct elements as introduced in the Method section. Annotations were initially provided independently by three annotators, followed by a consensus process to ensure consistency, semantic clarity, and alignment with real-world scientific procedures. An overview of the dataset structure and annotation scheme is shown in Fig.\ref{fig:pic_set}.
The evaluation metrics are introduced in the following section.

\section{EXPERIMENTS}

\subsection{Experimental Setup}

\begin{figure}[t]
  \centering
  \includegraphics[width=\linewidth]{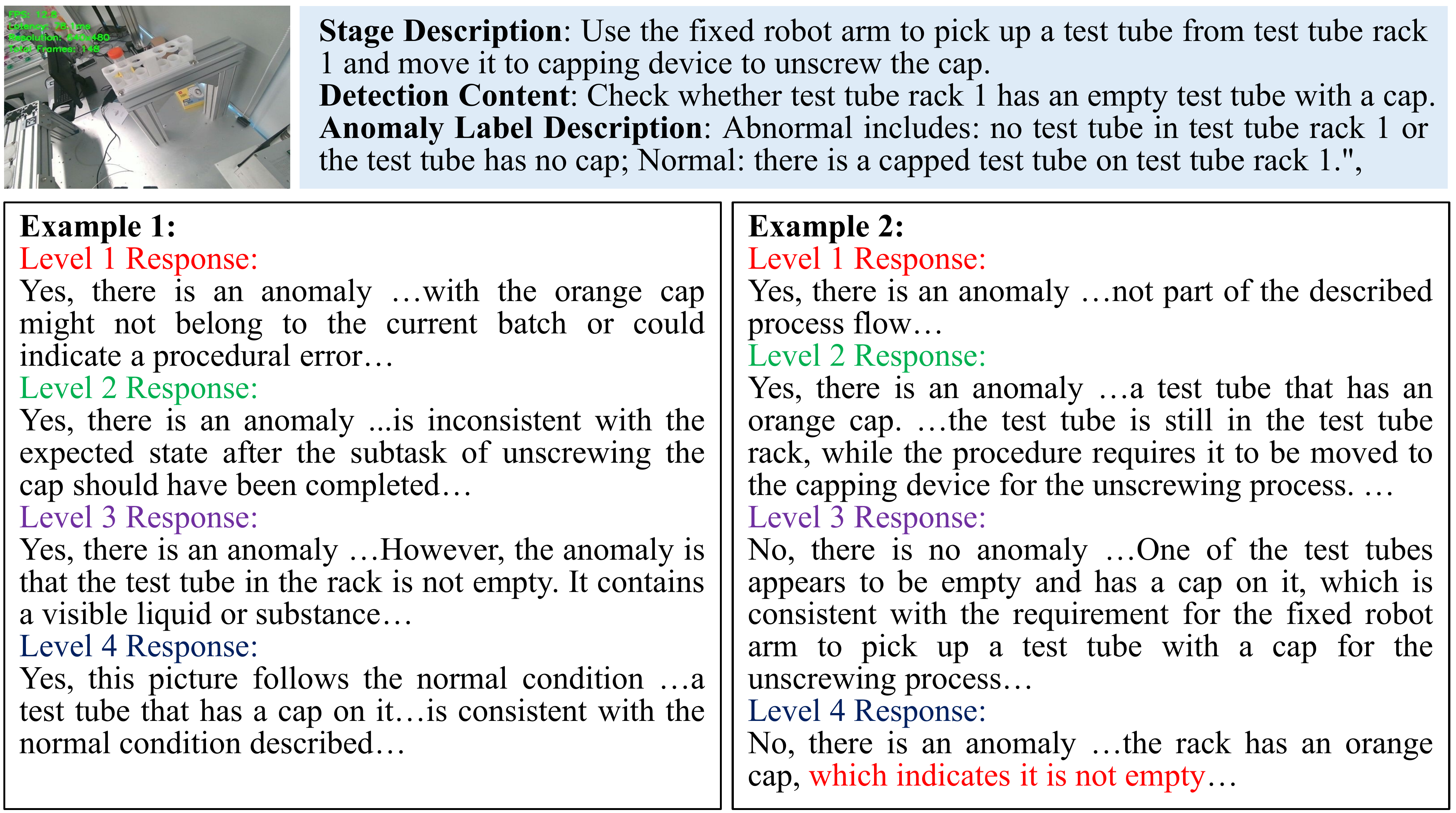}
  \caption{Qualitative examples illustrating the effects of hierarchical prompting. \textbf{Left:} The model corrects its focus and successfully detects the anomaly after receiving all four levels of prompts. \textbf{Right:} The model succeeds at Level 3 when provided with clear detection content, but fails at Level 4 due to misinterpretation of the anomaly label description, resulting in incorrect judgment.}
  \label{fig:examples}
\end{figure}

We evaluate two representative VLMs: GPT-4o and Qwen2.5-VL-72B-Instruct. The latter is abbreviated as QwenVL-72B. Both models are capable of performing image-text reasoning and support natural language prompting. Inference is conducted via publicly available APIs, without any fine-tuning.

In addition, we perform a real-world validation in a robotic laboratory, where anomaly detection is performed through a GPT-4o-driven reasoning pipeline to verify model behavior in actual experimental procedures.

\subsection{Evaluation Metrics}
To comprehensively evaluate model performance, we adopt the following four metrics:

\textbf{Accuracy (ACC)}: The percentage of samples correctly classified as normal or abnormal, indicating the overall effectiveness.

\textbf{False Positive Rate (FPR)}: The percentage of normal samples incorrectly predicted as abnormal. This reflects the model’s tendency to generate false alarms, which may cause unnecessary workflow interruptions.

\textbf{Missed Detection Rate (MDR)}: The percentage of abnormal samples mistakenly classified as normal, representing the model’s failure to detect true anomalies - a critical factor in high-risk experimental contexts.

\textbf{Uncertainty Rate (UR)}: The percentage of cases where the model cannot provide a confident judgment. This metric indicates the model’s ambiguity when handling visually complex or previously unseen samples, and provides insight into its robustness and reliability.

\subsection{Results and Analysis}

Table \ref{tab:metrics} reports the average performance of both models across the four prompt levels.

\begin{table}[hb]
\centering
\caption{Performance (\%) across different prompt levels.}
\label{tab:metrics}
\begin{tabular}{llcccc}
\toprule
\textbf{Model} & \textbf{Level} & \textbf{ACC$\uparrow$} & \textbf{FPR$\downarrow$} & \textbf{MDR$\downarrow$} & \textbf{UR$\downarrow$} \\
\midrule
\multirow{4}{*}{GPT-4o} 
    & Level 1 & 41.6 & 25.7 & 32.7 & 0.0 \\
    & Level 2 & 50.5 & 27.7 & 20.8 & 1.0 \\
    & Level 3 & 67.3 & 27.7 & 5.0  & 0.0 \\
    & Level 4 & 79.2 & 16.8 & 3.0  & 1.0 \\
\midrule
\multirow{4}{*}{QwenVL-72B} 
    & Level 1 & 41.6 & 38.6 & 19.8 & 0.0 \\
    & Level 2 & 49.5 & 23.8 & 26.7 & 0.0 \\
    & Level 3 & 57.4 & 19.8 & 22.8 & 0.0 \\
    & Level 4 & 61.4 & 35.6 & 3.0  & 0.0 \\
\bottomrule
\end{tabular}
\end{table}

\textbf{Overall Performance Trends}

\begin{itemize}
    \item GPT-4o shows a significant performance improvement, with accuracy increasing from 41.6\% at Level 1 to 79.2\% at Level 4, and MDR dropping to 3.0\%.
    \item QwenVL-72B demonstrates a more limited gain, with accuracy improving from 41.6\% to 61.4\%. A noticeable reduction in MDR (to 3.0\%) only occurs at Level 4, indicating its stronger reliance on explicit anomaly descriptions.
\end{itemize}

\textbf{Differences in False Positive Rate}

\begin{itemize}
    \item QwenVL-72B has a high FPR of 38.6\% at Level 1, suggesting a tendency to over-predict anomalies when contextual information is insufficient. Even with the most detailed prompts at Level 4, its FPR remains high at 35.6\%, reflecting sensitivity to prompt design.
    \item In contrast, GPT-4o maintains more a stable FPR across levels and achieves the lowest FPR of 16.8\% at Level 4, showing better robustness to prompt variations.
\end{itemize}

\textbf{Sensitivity to Prompt Granularity}

\begin{itemize}
    \item GPT-4o exhibits a sharp performance boost at Level 3, where MDR drops to 5.0\%, indicating that specifying the detection objective alone can significantly enhance its anomaly recognition.
    \item QwenVL-72B only shows notable improvement at Level 4, where explicit anomaly descriptions are included, suggesting that its reasoning capability is more dependent on semantic cues provided in the prompt.
\end{itemize}

In addition to quantitative metrics, we further conduct qualitative analysis to better understand how prompt granularity influences model decision-making. As shown in Fig.~\ref{fig:examples}, we present two representative cases. In the left example, the model initially fails due to inaccurate focus or object localization, but successfully identifies the anomaly after receiving all four levels of prompt information. In contrast, the right example remains misclassified at Levels 1 and 2, but correctly identifies the anomaly at Level 3 after receiving the detection content. However, when the anomaly label description is added at Level 4, the model incorrectly classifies the image as abnormal. This suggests that overly specific descriptions may inadvertently shift the model's attention toward secondary features, leading to misjudgment. These cases illustrate how hierarchical prompting can both enhance and, in some cases, misguide model reasoning, depending on how well the prompt aligns with the visual context.

\subsection{Real-World Demonstration}

To further validate the benchmark's applicability, we conducted a real-world test in a robotic chemical laboratory. The test focused on the operation where a robotic arm transfers a silicone bottle from the material table to the operation table.

\begin{figure}[t]
  \centering
  \includegraphics[width=\linewidth]{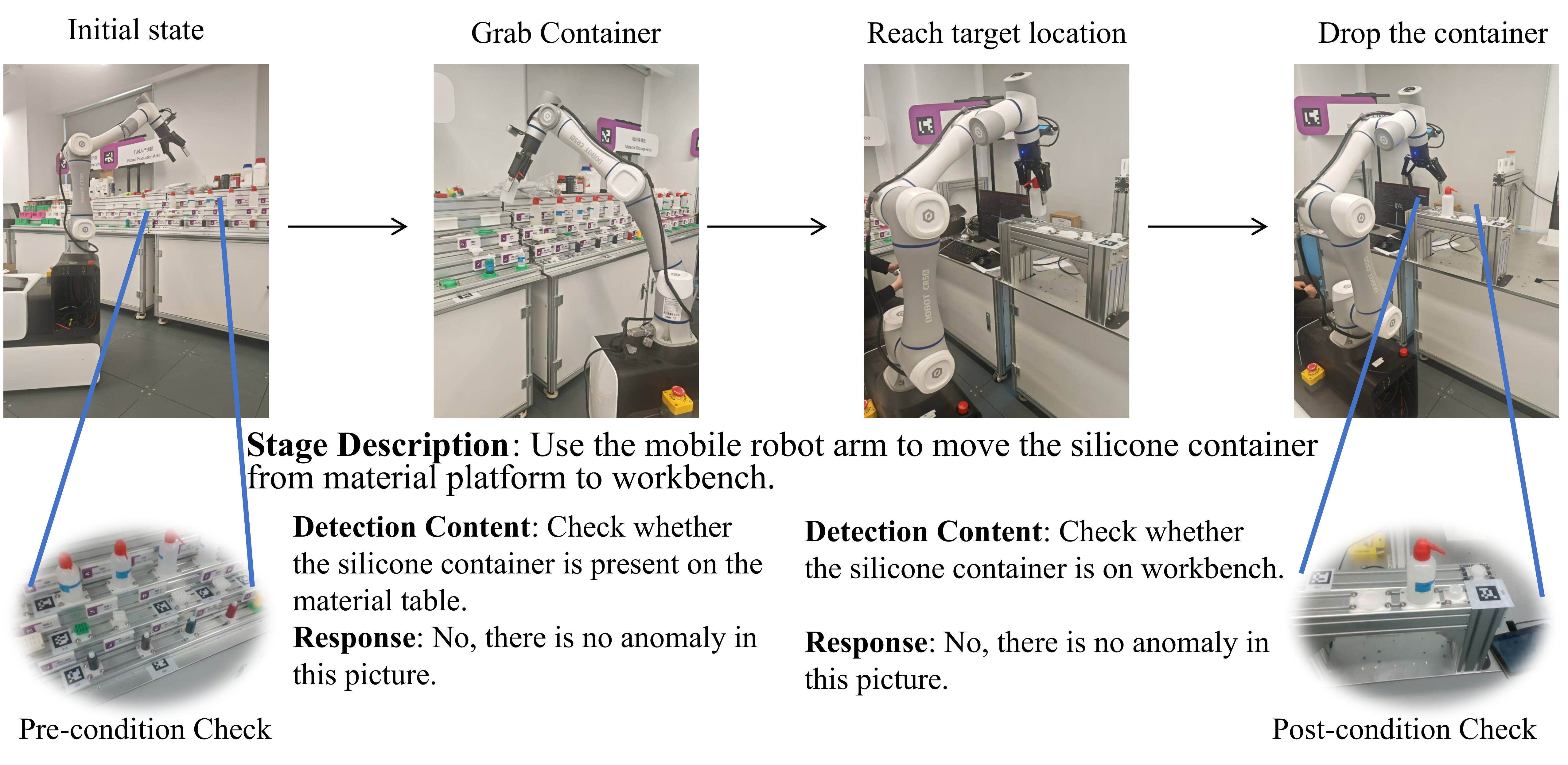}
  \caption{Real-world demonstration of anomaly detection: verifying the presence and correct placement of a silicone bottle during a robotic transfer operation.}
  \label{fig:real_demo}
\end{figure}

As shown in Fig.\ref{fig:real_demo}, the experiment is divided into two key checkpoints: before and after the execution of the transfer step. At the beginning of the step, the model receives the experiment background and step description as input, and is prompted to check whether the silicone bottle is present on the material table. The model correctly responds with no anomaly, confirming that the setup is valid.

After the robotic arm completes the transfer, another image is captured at the workbench. The model is again prompted to verify whether the silicone bottle has been successfully placed. The result is also no anomaly, demonstrating the model's ability to monitor the correctness of step execution under real-world conditions.

\section{CONCLUSIONS}

This paper proposes a vision-language reasoning approach for process anomaly detection in scientific experiments, leveraging VLMs guided by hierarchical prompts and CoT inference for step-wise anomaly judgment. To support systematic evaluation, we construct a visual reasoning benchmark based on a real-world chemical workflow. Experimental results show that prompt granularity significantly affects model performance. Real-world validation confirms the method’s effectiveness in detecting execution anomalies from first-person visual input. Future work will expand the benchmark and explore automatic prompt generation for broader applicability, and generate descriptive anomaly reports from VLM reasoning traces to support more explainable and actionable outputs.

\addtolength{\textheight}{-12cm}   







\bibliographystyle{IEEEtran} 
\bibliography{references} 

\end{document}